\documentclass{article}
\usepackage{spconf}
\usepackage[bookmarks=false,hyperfigures=false,hidelinks=true]{hyperref}

\makeatletter
\def\input@path{{Plots/}}
\makeatother
\usepackage{verbatim}
\usepackage{relsize} %
\usepackage{amsmath}

\usepackage{amssymb, subfigure, url, doi, xspace}
\usepackage{mathtools}
\usepackage[linesnumbered,algoruled,boxed,lined]{algorithm2e}

\usepackage{ifpdf}
\ifpdf
  \usepackage[dvips, final]{graphicx}
  \DeclareGraphicsExtensions{.pdf}
\else
  \usepackage[ final]{graphicx}
  \DeclareGraphicsExtensions{.pstex,.ps,.eps}
\fi
\usepackage{epstopdf}
\usepackage{color}

\newcommand{\ie}{i.e.\xspace}

\newcommand{\mb}[1]{\mathbf{#1}}

\providecommand{\norm}[1]{\left\lVert#1\right\rVert}
\DeclarePairedDelimiterX{\normsz}[1]{\lVert}{\rVert}{#1}

\DeclareMathOperator*{\argmin}{arg\,min}

\newcommand{\resizemath}[2]{%
  \resizebox{#1}{!}{%
    $ \displaystyle #2 $%
  }%
}%
\def \co {\mathcal{O}}

\def \F  {\mathcal{F}}

\date{\today}
\begin{document}
\title{Online Convolutional Dictionary Learning}
\name{Jialin Liu$^{a}$ \qquad Cristina Garcia-Cardona$^{b}$ \qquad
Brendt Wohlberg$^{c}$\thanks{This research was supported by the
    U.S. Department of Energy via the LANL/LDRD Program.} \qquad Wotao Yin$^{a}$}
\address{$^{a}$Department of Mathematics, UCLA, Los Angeles, CA\\
         $^{b}$CCS Division, Los Alamos National Laboratory, Los
         Alamos, NM\\
         $^{c}$Theoretical Division, Los Alamos National Laboratory,
         Los Alamos, NM}

\maketitle

\begin{abstract}
 While a number of different algorithms have recently been proposed for convolutional dictionary learning, this remains an expensive problem. The single biggest impediment to learning from large training sets is the memory requirements, which grow at least linearly with the size of the training set since all existing methods are batch algorithms. The work reported here addresses this limitation by extending online dictionary learning ideas to the convolutional context.
\end{abstract}


\section{Introduction}
\label{sec:intro}

Sparse representations and dictionary learning have become ubiquitous techniques in signal and image processing, computer vision, and machine learning~\cite{mairal-2014-sparse}. Most dictionary learning algorithms (e.g.~\cite{aharon-2006-algorithm, engan-1999-method}) are \emph{batch} methods in that they require the access to all training data when they start, so the training data size is limited by the amount of available memory. \emph{Online} methods~\cite{skretting-2010-recursive, mairal-2010-online, lu-2013-robust, naderahmadian-2016-gaw-rls}, in contrast, are designed to operate on small subsets of the training data at a time, making it possible to process a large training data set with limited memory.  These methods continuously aggregate past training data, updating the current learned dictionary to incorporate the sparse codes obtained for the new training data. The updates depend on accumulating the sparse codes computed while training, and do not require accessing previous sparse codes. These methods can therefore run in constant memory and at a computation cost linear in the total training data size.

Consider the linear representation, $D \mb{x} \approx \mb{s}$, where $\mb{s}$ is a signal of size $N$ to represent, $D$ is a dictionary, and $\mb{x}$ is the representation. A \emph{convolutional representation}~\cite{zeiler-2010-deconvolutional} implements $D\mb{x}$ as a sum of convolutions, $\sum_{m=1}^M \mb{d}_m \ast \mb{x}_m \approx \mb{s}$, where $\mb{d}_m$ are dictionary filters, and the representation $\{\mb{x}_m\}_{m=1}^M$ is a set of coefficient maps, each map $\mb{x}_m$ having the same size $N$\ as the signal $\mb{s}$.  Given $\{\mb{d}_m\}$ and $\mb{s}$, a \emph{sparse} convolutional representation can be obtained by solving the Convolutional Basis Pursuit DeNoising (CBPDN) $\ell_1$-minimization problem \vspace{-1mm}
\begin{align}
  \argmin_{\{\mb{x}_{m}\}_{m=1}^M} \frac{1}{2}
  \normsz[\Big]{\sum_m \mb{d}_m \ast \mb{x}_{m} - \mb{s}}_2^2 +
  \lambda  \sum_m \norm{\mb{x}_{m}}_1  \; .
\label{eq:cbpdn}
\vspace{-1mm}
\end{align}
Given a set of $K$ training signals $\{\mb{s}_k\}_{k=1}^K$, the dictionary learning problem is \vspace{-1mm}
\begin{align}
\label{training}
\resizemath{.9\hsize}{
  \argmin_{\{\mb{d}_{m}\},\{\mb{x}_{k,m}\}} \frac{1}{2} \sum_k
  \normsz[\Big]{\sum_m \mb{d}_m \ast \mb{x}_{k,m} - \mb{s}_k}_2^2 +
  \lambda \sum_k \sum_m \norm{\mb{x}_{k,m}}_1  } \nonumber \\
 \resizemath{.65\hsize}{ \text{ subject to}  \norm{\mb{d}_{m}}_2 = 1,
\; \forall m \in \{1,\ldots,M\} \; ,}
\vspace{-1mm}
\end{align}
where the coefficient maps ${\mb{x}_{k,m}}$, $m \in \{1,\ldots,M\}$, represent $\mb{s}_k$, $k \in \{1,\ldots,K\}$. The norm constraint avoids the scaling ambiguity between $\mb{d}_m$ and $\mb{x}_{k,m}$. As the notation is cumbersome, it is convenient to define the linear operator $D_m$ such that $D_m \mb{x}_{k,m} = \mb{d}_m \ast \mb{x}_{k,m}$ and let $D \triangleq \big(D_1~D_2\cdots D_M\big) $ and $\mb{x}_k \triangleq \big(\mb{x}_{k,1}^T~\mb{x}_{k,2}^T\cdots \mb{x}_{k,M}^T\big)^T$. Then, we can write $D \mb{x}_k = \sum_{m=1}^M \mb{d}_m \ast \mb{x}_{k,m} \approx \mb{s}_k$.

The most recent approaches for solving \eqref{training}, all of which are \emph{batch} methods, use alternating minimization over $\{\mb{x}_{k,m}\}$ and $\{\mb{d}_{m}\}$, where each minimization subproblem is approximated by performing a few iterations of ADMM~\cite{bristow-2013-fast, heide-2015-fast, wohlberg-2016-efficient, sorel-2016-fast, wohlberg-2016-convolutional, wohlberg-2016-boundary}.  When $K$ is large, the $\mb{d}_m$ update subproblem is computationally expensive since it depends on all the $KM$ maps $\mb{x}_{k,m}$ of size $N$ each, thus preventing the use of a large training set.  The purpose of the present work is to develop online dictionary learning methods for training data sets that are much larger than those that are presently feasible.

\section{Online Dictionary Learning}
\label{sec:olcdl}

In the online setting, training signals are processed in a streaming fashion, $\mb{s}^{(1)}, \mb{s}^{(2)}, \cdots, \mb{s}^{(t)},\cdots$.  The coefficient maps $\{\mb{x}^{(t)}_m\}_{m=1}^M$ of the $t^{\mathrm{th}}$ training signal $\mb{s}^{(t)}$ are computed via CBPDN (\ref{eq:cbpdn}) using the latest dictionary $\{\mb{d}_m^{(t-1)}\}_{m=1}^M$, \vspace{-1mm}
\begin{equation}
\label{cbpdn}
\{\mb{x}^{(t)}_m\}_{m=1}^M \leftarrow \text{CBPDN}\Big(\mb{s}^{(t)}, \{\mb{d}_m^{(t-1)}\}_{m=1}^M\Big) \;,
\vspace{-1mm}
\end{equation}
which is a relatively cheap step since only the current signal $\mb{s}^{(t)}$ is involved. Define the loss function for $\mb{s}^{(t)}$ \vspace{-1mm}
\begin{equation}
\label{eq:loss}
f^{(t)}(D)\triangleq \frac{1}{2} \Big\| \sum_{m} \mb{x}^{(t)}_{m} * \mb{d}_m - \mb{s}^{(t)} \Big\|_2^2 \;.
\vspace{-1mm}
\end{equation}
The simplest way to update the dictionary is to minimize the loss function at the current iterate, \vspace{-1mm}
\begin{equation}
\label{eq:naive}
D^{(t)} \leftarrow \argmin_{D} f^{(t)}(D) + \iota_{\mathrm{CPN}}(D) \;,
\vspace{-1mm}
\end{equation}
where $\iota_{\mathrm{CPN}}$ is the indicator function of the constraint set for filter support and normalisation~\cite{wohlberg-2016-efficient}.  However, this approach may overfit each $\mb{s}^{(t)}$ and never converge to a dictionary that represents the features of the entire training sequence.  A better approach, inspired by \cite{mairal-2010-online}, introduces the \emph{surrogate function}
\vspace{-1mm}
\begin{equation}
\label{eq:surrogate}
\F^{(t)}(D) = f^{(1)}(D) + \cdots + f^{(t)}(D) \;,
\vspace{-1mm}
\end{equation}
based on which, the updated dictionary is computed as,
\vspace{-1mm}
\begin{equation}
\label{d-update}
D^{(t)} \leftarrow \argmin_{D} \F^{(t)}(D) + \iota_{\mathrm{CPN}}(D) \;.
\vspace{-1mm}
\end{equation}
Modified variants of (\ref{d-update}) are proposed in Section \ref{sec:modify} below, with its fast algorithm presented in Section \ref{sec:fista}.

\subsection{Acceleration via Modified Surrogate Function}
\label{sec:modify}

At the current time $t$, the dictionary is the result of an accumulation of past coefficient maps $\mb{x}^{(\tau)}_m$, $\tau < t$ which were computed with the then-available dictionaries. A way to balance accumulated past contributions and information provided by the new training samples is to compute a weighted combination of these contributions, as routinely done in other online schemes~\cite{skretting-2010-recursive, mairal-2010-online}. This combination considers more strongly the more recent updates, since those are the result of a more extensively trained dictionary.

Consider the surrogate function (\ref{eq:surrogate}) written recursively as
\vspace{-1mm}
\begin{equation}
\F^{(t)}(D) = \F^{(t-1)}(D) + f^{(t)}(D) \;,
\vspace{-1mm}
\end{equation}
and instead of a direct combination, use a factor to weight past (\ie outdated) contributions
\vspace{-1mm}
\begin{equation}
\label{surrogate-f}
\F^{(t)}_{\text{mod}} (D)= \alpha^{(t)} \F^{(t-1)}_{\text{mod}}(D) + f^{(t)}(D) \;.
\vspace{-1mm}
\end{equation}
Here $\alpha^{(t)}\in (0,1)$ is a \emph{forgetting factor}, which has its own time evolution~\cite{mairal-2010-online}
\vspace{-1mm}
\begin{equation}
\label{forget}
\alpha^{(t)} = (1-1/t)^p
\vspace{-1mm}
\end{equation}
regulated by the \emph{forgetting exponent} $p>0$.  This is a reasonable choice since, as $t$ increases, the factor $\alpha^{(t)}$ increases ($\alpha^{(t)} \to 1$ as $t \to \infty$), reflecting the increasing accuracy of the past information as the training progresses.

A large forgetting factor $\alpha$ (small $p$) can be expected to lead to a stable algorithm since all the training signals are given nearly equal weights as their information is accumulated in $D$. However, it also leads to slow convergence.  An extreme case is as $p \to 0$, $\alpha^{(t)} \to 1$, which recovers (\ref{eq:surrogate}). A small forgetting factor, conversely, leads to faster convergence since it gives past, less accurate information lower weights. But if the factor is too small ($p$ is too large), the surrogate function is overwhelmingly influenced by the current training signal $\mb{s}^{(t)}$, causing the convergence to be unstable. As $p \to \infty$, we have $\alpha^{(t)} \to 0$, so only the loss function $f^{(t)}(D)$ of the current $\mb{s}^{(t)}$ is considered, which recovers (\ref{eq:naive}).
Based on the modified surrogate function $\F^{(t)}_{\text{mod}} (D)$ in \eqref{surrogate-f}, the dictionary update  (\ref{d-update}) is  modified correspondingly to
\vspace{-2mm}
\begin{equation}
\label{d-update-modi}
D^{(t)} \leftarrow \argmin_{D} \F^{(t)}_{\text{mod}}(D) + \iota_{\mathrm{CPN}}(D) \;.
\vspace{-2mm}
\end{equation}

\begin{algorithm}[t]
\SetKwInOut{initial}{Initiate}
\initial{Initialize $D^{(0)}$ with random dictionary.\\$\hat{A}^{(0)}_{\text{mod}} \leftarrow 0, \hat{\mb{b}}^{(0)}_{\text{mod}} \leftarrow 0.$}
\For{$t=0,1,\cdots, T$} {
Sample a signal $\mb{s}^{(t)}$.\\
Solve sparse coding problem (\ref{cbpdn}).

Compute FFT: $\hat{\mb{x}}^{(t)} = \text{FFT2}(\mb{x}^{(t)})$. \\
Accumulate with forgetting factor $\alpha^{(t)}$,
 \vspace{-2mm}
\begin{equation}
\label{accumulating}
\begin{aligned}
\hat{A}^{(t)}_{\text{mod}} \leftarrow \alpha^{(t)} \hat{A}^{(t-1)}_{\text{mod}} + (\hat{\mb{x}}^{(t)})^H (\hat{\mb{x}}^{(t)} )\\
\hat{\mb{b}}^{(t)}_{\text{mod}} \leftarrow \alpha^{(t)} \hat{\mb{b}}^{(t-1)}_{\text{mod}} + (\hat{\mb{x}}^{(t)})^H (\hat{\mb{s}}^{(t)})
\end{aligned}
\vspace{-2mm}
\end{equation}\\
$D$-update:  Solve (\ref{d-update-modi}) via Algorithm \ref{algo-fista}.
}
\KwOut{$D^{(T)}$}
\caption{Main Algorithm: Online Convolutional Dictionary Learning}\label{online-cdl}
\end{algorithm}

\subsection{Minimizing Modified Surrogate Function}
\label{sec:fista}

A popular approach for solving quadratic minimization problems like (\ref{d-update}) and (\ref{d-update-modi}) is Fast Iterative Shrinkage-Thresholding (FISTA)~\cite{beck2009fast}, which computes a gradient at each step. According to the notation in Section \ref{sec:intro}, each loss function can be written as $f^{(t)}(D) = \frac{1}{2} \|D \mb{x}^{(t)} - \mb{s}^{(t)}\|^2_2$. Thus, the gradient for the surrogate function can be computed as \vspace{-2mm}
\begin{equation}
\resizemath{.88\hsize}{
\nabla \F^{(t)}(D) = \Big(\sum_{\tau=1}^t (\mb{x}^{(\tau)})^T \mb{x}^{(\tau)}\Big) D - \Big(\sum_{\tau=1}^t (\mb{x}^{(\tau)})^T \mb{s}^{(\tau)}\Big) \;.}
\vspace{-2mm}
\end{equation}
We cannot follow this formula directly since the cost would increase linearly in $t$.  Instead we perform the iterative updates~\cite{mairal-2010-online} \vspace{-1mm}
\[
\resizemath{.9\hsize}{
A^{(t)} = A^{(t-1)} + (\mb{x}^{(t)})^T \mb{x}^{(t)} \;, \quad \mb{b}^{(t)} = \mb{b}^{(t-1)} + (\mb{x}^{(t)})^T \mb{s}^{(t)} \;,}
\vspace{-1mm}
\]
at a constant cost. So, at a constant cost, we can also compute $\nabla \F^{(t)}(D) = A^{(t)}D - \mb{b}^{(t)}$.

A further significant improvement is to take advantage of the convolutional property of the linear operator $D$.  The convolution is implemented in the frequency domain, using the Fast Fourier Transform (FFT). Inspired by the frequency domain FISTA variant for the CBPDN problem~\cite{wohlberg-2016-efficient}, we propose a frequency domain FISTA to solve (\ref{d-update-modi}), which is described in Algorithm \ref{algo-fista}.  Each loss function in the frequency domain has the form \vspace{-1mm}
\begin{equation}
\resizemath{.88\hsize}{
{f}^{(t)}(\hat{D}) \triangleq \frac{1}{2} \Big\| \hat{D} \hat{\mb{x}}^{(t)} - \hat{\mb{s}}^{(t)} \Big\|_2^2 = \frac{1}{2} \Big\| \sum_{m} \hat{\mb{d}}_m \odot \hat{\mb{x}}^{(t)}_m - \hat{\mb{s}}^{(t)} \Big\|_2^2 \;,
}
\vspace{-1mm}
\end{equation}
where $\hat{\cdot}$ denote frequency-domain values, $\hat{\mb{x}}^{(t)}$ is obtained by applying the Fourier transforms to each $\mb{x}_m^{(t)}$ in $\mb{x}^{(t)}$, and ``$\odot$'' is point-wise multiplication. Then, the gradient can be computed as $\nabla {\F}^{(t)}(\hat{D}) = \hat{A}^{(t)}\hat{D} - \hat{\mb{b}}^{(t)}$, where $\hat{A}^{(t)}$ and $\hat{\mb{b}}^{(t)}$ are iteratively updated through \vspace{-1mm}
\[
\hat{A}^{(t)} = \hat{A}^{(t-1)} + (\hat{\mb{x}}^{(t)})^H \hat{\mb{x}}^{(t)}\;, \quad \hat{\mb{b}}^{(t)} = \hat{\mb{b}}^{(t-1)} + (\hat{\mb{x}}^{(t)})^H \hat{\mb{s}}^{(t)}.
\vspace{-1mm}
\]
This accumulation is weighted by the forgetting factor, as derived from the modified surrogate function, to yield expression (\ref{accumulating}).  The main algorithm is given in Algorithm \ref{online-cdl}. Its $D$-update step calls \textit{frequency domain FISTA} which is described in Algorithm~\ref{algo-fista}.

Direct extension of the online approach for regular dictionary learning leads to matrices $A^{(t)}$ and $\hat{A}^{(t)}$ of size $\co(M^2N^2)$, which is prohibitive except for very small $N$ ($M$ is usually much smaller than $N$, so the quadratic order is less problematic). However, since the frequency-domain product $(\hat{\mb{x}}^{(t)})^H \hat{\mb{x}}^{(t)}$ has only $\co(M^2N)$ non-zero values, this structure can be exploited to obtain a corresponding reduction in storage requirements for $\hat{A}^{(t)}$.

\begin{algorithm}[t]
\SetKwInOut{initial}{Initiate}
\KwIn{Information matrix $\hat{A}^{(t)}_{\text{mod}}$ and $\hat{\mb{b}}^{(t)}_{\text{mod}}$.}
\initial{Let $G^0 = D^{(t-1)}, \hat{G}^0 = \text{FFT2}(G^0)$.
\\Auxiliary variable $G^0_{\text{aux}} = G^0$. \\Let $\gamma^0 = 1$ for acceleration.}
\For{$j = 0,1,2,\ldots$ until convergence} {
Compute FFT, $\hat{G}^j_{\text{aux}}= \text{FFT2}(G^j_{\text{aux}})$. \\
Compute gradient in frequency domain,
\vspace{-2mm}
\begin{equation}
\label{fre-gra}
\nabla {\F}^{(t)}_{\text{mod}}(\hat{G}_{\text{aux}}^{j}) = \hat{A}^{(t)}_{\text{mod}} \hat{G}_{\text{aux}}^j - \hat{\mb{b}}^{(t)}_{\text{mod}} \;.
\vspace{-2mm}
\end{equation}\\
Compute dictionary,
\vspace{-2mm}
\begin{equation}
\label{ista}
\resizemath{.86\hsize}{
G^{j+1} = \text{proj}_{C_{PN}} \bigg(\text{IFFT2}\Big(\hat{G}^{j}_{\text{aux}} - \eta \nabla {\F}^{(t)}_{\text{mod}} (\hat{G}^{j}_{\text{aux}})\Big)\bigg) \;.
}
\vspace{-2mm}
\end{equation}
Compute auxiliary dictionary (Nesterov acceleration) for next step,
\vspace{-2mm}
\[
\gamma^{j+1} = \Big(1+\sqrt{1+4(\gamma^j)^2}\Big) / 2 \;,
\vspace{-2mm}
\]
\vspace{-2mm}
\begin{equation}
\label{nestrov}
G^{j+1}_{\text{aux}} = G^{j+1} + \frac{\gamma^{j}-1}{\gamma^{j+1}} (G^{j+1}-G^j) \;.
\vspace{-2mm}
\end{equation}
}
\KwOut{$D^{(t)} \leftarrow G^J$, where $J$ is the last iterate.}
\caption{D-update: Frequency domain FISTA for solving (\ref{d-update-modi})}\label{algo-fista}
\end{algorithm}

\subsection{Region Sampling (Limited Memory Version)}
\label{sec-samp}

In Section \ref{sec:fista}, the information matrix $\hat{A}^{(t)}_{\text{mod}}$ of size $\co(M^2N)$ is maintained and updated. (Recall that $M$ is the total number of dictionary filters and $N$ is the signal dimension.) When $M$ and $N$ are large, $\hat{A}^{(t)}_{\text{mod}}$ requires a large amount of memory.  To reduce memory size, we sample small regions of the whole signal. Specifically, given a current signal $\mb{s}^{(t)} \in N$, we sample small regions ${\mb{s}^{(t)}_{\text{samp},1}, \mb{s}^{(t)}_{\text{samp},2}, ... } \in \tilde{N}$, with $\tilde{N} < N$, and treat them as if they were different signals.

In this way, the training signal sequence becomes:
\vspace{-2mm}
\[
\{\mb{s}^{(t)}_{\text{samp}}\}_{t} \triangleq \{ \mb{s}^{(1)}_{\text{samp},1}, \cdots,\mb{s}^{(1)}_{\text{samp},n}, \mb{s}^{(2)}_{\text{samp},1}, \cdots, \mb{s}^{(2)}_{\text{samp},n}, \cdots\} \;.
\vspace{-1mm}
\]
In our experiments, we sample each $256 \times 256$ image to obtain small $64 \times 64$ regions before Algorithm \ref{online-cdl} is called. For the experiments reported here we use circular boundary conditions rather than the more careful boundary handling~\cite{heide-2015-fast, wohlberg-2016-boundary} that would be necessary for smaller regions. We call this approach ``Online-Samp.''.

\section{Results}
\label{sec:rslt}

All the experiments are conducted using MATLAB R2016a running on a workstation with 2 Intel Xeon(R) X5650 CPUs clocked at 2.67GHz.  The dictionary size is $8 \times 8 \times 32$ and the training and testing image size is $256 \times 256$.  As in~\cite{mairal-2010-online}, dictionaries are evaluated by comparing the functional values obtained by computing CBPDN (\ref{eq:cbpdn}) on the test set. The training set consists of 50 images and the test set consists of 5 separate images. Four of the training set images were standard images (Lena, Barbara, Kiel, Mandrill), and the remainder of the training images, and the testing images, were cropped and rescaled from a set of images, of a variety of scenes, obtained from Flickr.

\subsection{Effect of Forgetting Exponent $p$}
\label{sec:p-results}

\begin{figure}
\centering
\small
\subfigure[$1 \leq p \leq 10$: larger $p$ leads to faster convergence.]{
\label{fig_1_1}
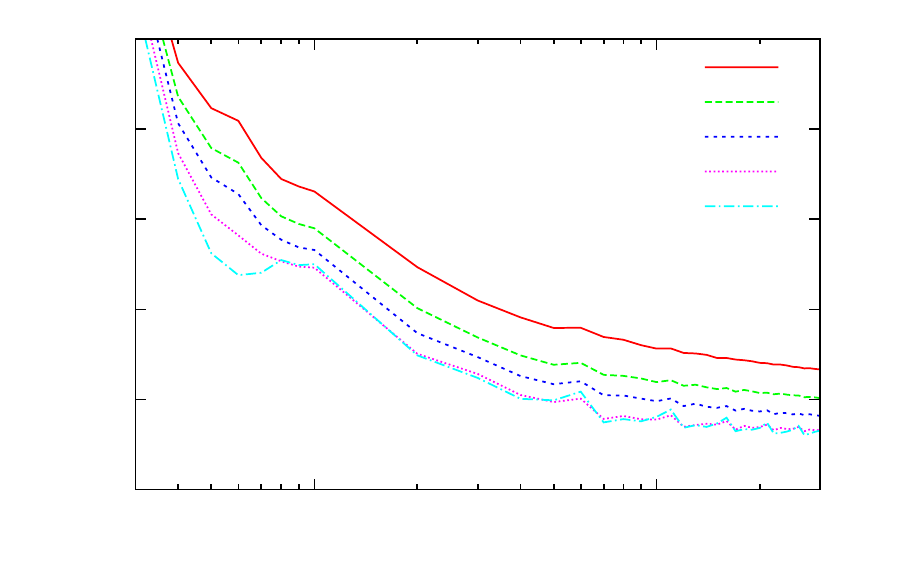}
\subfigure[$10 \leq p \leq \infty$: too large $p$ leads to instability.]{
\label{fig_1_2}
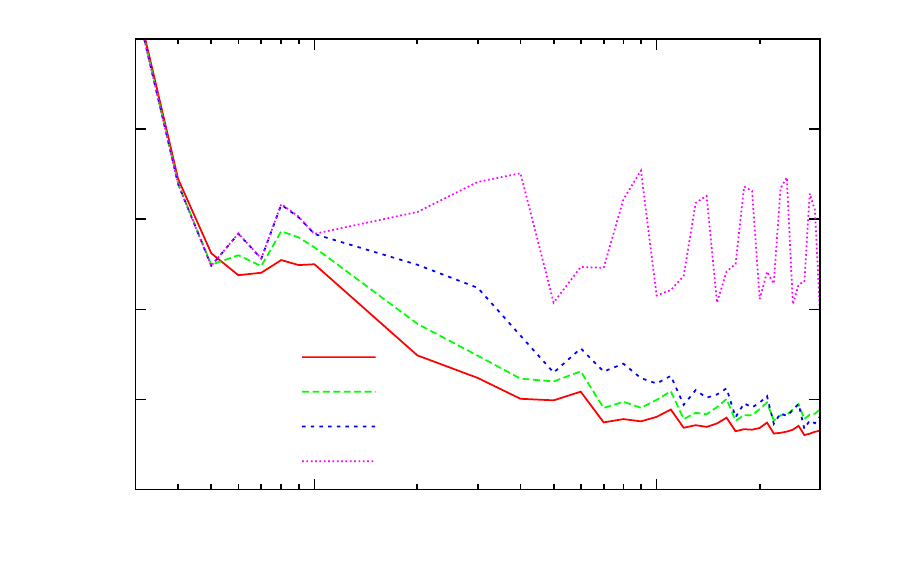}
\caption{
A comparison of the convergence behaviour of the online dictionary learning algorithm for different forgetting exponents $p$. Note that the functional value is evaluated on the testing set, not the training set.
}
\label{fig:forget}
\end{figure}
An efficient algorithm requires a good choice of the forgetting factor $\alpha^{(t)}$.  In this paper, we used the forgetting factor evolution defined in (\ref{forget}), where $p$ is the crucial exponent parameter to tune, as we discussed in Section \ref{sec:modify}. The experiments reported in Fig \ref{fig:forget} compare the convergence resulting from difference choices of $p$ for the full training set of 50 images.  In Fig \ref{fig_1_1}, when $1 \leq p \leq 10$, the algorithm is faster when $p$ increases ($p=5$ is almost the same with $p=10$). In Fig \ref{fig_1_2}, if $p$ increases continuously, the algorithm gets more and more unstable (for $p=\infty$, the algorithm reduces to the ``naive'' update scheme (\ref{eq:naive})). Thus, $p=5$ is a reasonable choice.

\subsection{Comparison with Batch Learning}

\begin{figure}
\centering \small
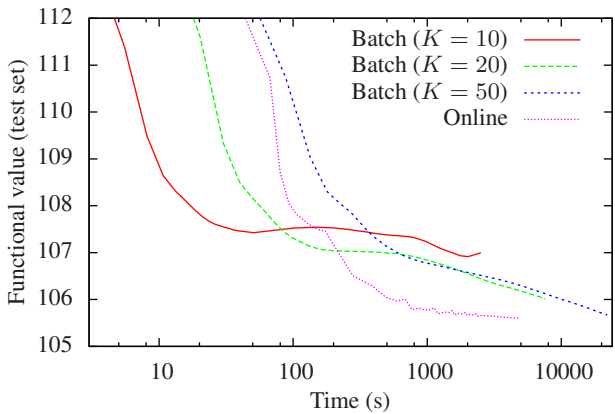
\caption{Computation time comparison of batch and online learning. Note that the functional value is evaluated on the testing set, not the training set.
}
\label{fig:method}
\end{figure}

In this section, online convolutional dictionary learning with a set of 50 training images and with parameter $p=5$ is compared with batch learning~\cite{wohlberg-2016-efficient, wohlberg-2016-sporco} with training sizes of 10, 20, and 50. The result is shown in Fig \ref{fig:method}. According to this figure, online convolutional dictionary learning is the best after 200 seconds.  For batch learning, a small training set leads to inaccurate results but a large training set leads to a large computational cost. Online learning handles only one image at a time, and accumulates the previous information in a compact way, making it efficient for a large training set.

\subsection{Region Sampling (Limited Memory Version)}
\label{sec:samp-results}

\begin{figure}
\centering \small
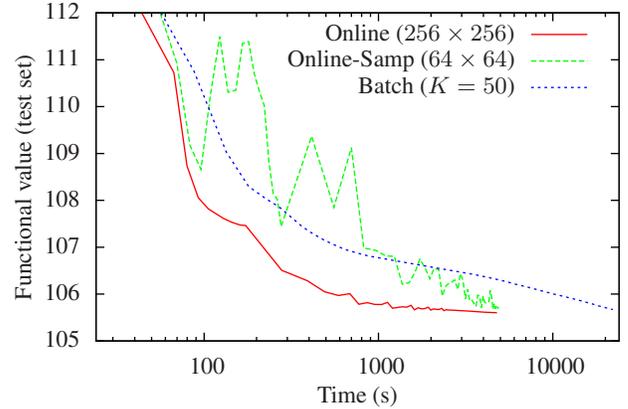
\caption{A comparison of Online, Online-samp, and Batch algorithms with a training set of 50 images. The functional value is evaluated on the testing set, not the training set.}
\label{fig:samp2}
\end{figure}

To avoid high memory usage, we use the technique proposed in Section \ref{sec-samp}, and consider a sample region with size $64 \times 64$.  We found experimentally that $p=40$ is a good choice for size $64 \times 64$.  As shown in Fig. \ref{fig:samp2}, the Online-Samp. scheme is not as stable as online scheme without sampling, but it still shows good performance after enough iterations, and requires substantially less memory, as shown in Table \ref{tab:memory}.

\begin{table}
\centering
\begin{tabular}{|c|r|}
\hline
Schemes & Memory (MB) \\
\hline
Batch ($K=10$) & 618\\
\hline
Batch ($K=20$) & 1197\\
\hline
Batch ($K=50$) & 2902 \\
\hline
Online ($256 \times 256$)& 1213 \\
\hline
Online-Samp. ($64 \times 64$) & 133 \\
\hline
\end{tabular}
\caption{Memory Usage Comparison in Megabytes.
}
\label{tab:memory}
\end{table}

\section{Conclusions}
\label{sec:concl}

We have proposed the first online convolutional dictionary learning algorithms capable of learning from a training image set of arbitrary size. Our approaches are based on an extension of ideas from online dictionary learning for standard sparse representations. The first of these processes an entire training image at a time; while the $\co(NM^2)$ memory cost is vastly better than the $\co(N^2M^2)$ cost that would correspond to a direct extension of prior methods for standard sparse representations, it can still be high when $M$ is large. The second approach further reduces memory usage by sampling regions from each training image, at the expense of somewhat worse convergence behaviour.

\bibliographystyle{IEEEtranD}
\bibliography{online}
\end{document}